\documentclass{article}
\usepackage[utf8]{inputenc}

\usepackage{graphicx}
\usepackage{algorithm}

\usepackage{algpseudocode}
\usepackage{authblk}
\usepackage{caption}

\usepackage{subfig}
\usepackage{fancyhdr}
\usepackage[numbers]{natbib}
\usepackage{fancyhdr}
\usepackage{graphicx}
\usepackage{algorithm}
\usepackage{algpseudocode}
\usepackage{subfig}

\title{Adaptive Neural Networks using Residual Fitting}
\author[1]{
  Noah Ford (\texttt{Noah.Ford@jhuapl.edu}) }
\author[1]{John Winder }
\author[1]{Joshua McClellan }
\affil[1]{  Johns Hopkins University Applied Physics Laboratory,\\
	11100 Johns Hopkins Road, Laurel, Maryland 20723, USA\\}

\begin{document}

\maketitle
\thispagestyle{fancy}

\section{Abstract}
Current methods for estimating the required neural-network size for a given problem class have focused on methods that can be computationally intensive, such as neural-architecture search and pruning. In contrast, methods that add capacity to neural networks as needed may provide similar results to architecture search and pruning, but do not require as much computation to find an appropriate network size. Here, we present a network-growth method that searches for explainable error in the network's residuals and grows the network if sufficient error is detected. We demonstrate this method using examples from classification, imitation learning, and reinforcement learning. Within these tasks, the growing network can often achieve better performance than small networks that do not grow, and similar performance to networks that begin much larger.

\section{Introduction}
Prior to training a neural network on a set of data, it is difficult to estimate the size or shape of the network required to fit the complexity of the data. The network's ability to fit a data set is also dependent on the network's initialization, which further complicates the search for the ``right" network size. For most problems, a smaller network than typically used would suffice if we only knew how to initialize the weights \cite{frankle2019lottery}. Since we have little a priori knowledge of the structure of the best-performing networks, we can either train a network that is much larger than needed, or we can search for an appropriate smaller network using techniques such as neural architecture search \cite{elsken2019neural}, network pruning \cite{blalock2020state}, and network growing. In this paper, we present and test a new technique for networking growing that trains a small, neural network on the residuals of a larger network. This method can be used in addition to, and as a replacement for, neural architecture search, network pruning, and other growth methods.

General architecture searches can find network sizes and shapes that are appropriate for a set of problems with a given complexity \cite{alparslan2021searching}. These methods train many neural networks, either simultaneously or in sequence, to discover the right architecture for a given problem. For example, \cite{so2019evolved} uses neural architecture search to find a more efficient and performant transformer-like network for translation tasks. While neural-architecture search can produce more efficient networks, the search itself is computationally intensive, as it necessitates the training of many networks.

Network pruning is another method to find suitable network structures, but often requires less computational cost than architecture searches. Pruning is particularly popular for finding small, performant networks for fast inference. Several criteria have been proposed for choosing which nodes and at what time to prune (\cite{molchanov2017pruning}, \cite{blalock2020state}). These methods can help find performant, smaller networks, but they often require training much larger networks to find the most performant subnetwork (\cite{frankle2019lottery}). Furthermore, pruning methods may produce small, neural networks that perform well on a particular problem, but do not adapt well to changes in data.

There has been less research into network growth methods than into architecture search or pruning, but network growth has benefits such as allowing for smaller initial networks and fitting distribution shifts in the data with additional growth during training. To grow a neural network, we must decide: 1. when to grow and 2. how to grow. To help determine when to grow a neural network, the method presented in \cite{kilcher2019escaping} grows a neural network when the loss function is experiencing a plateau to help escape gradient-descent slow-downs. To determine how to grow, the method in \cite{evci2022gradmax} presents Gradient Maximizing Growth to maximize the gradients of new weights to accelerate training. Focusing on non-stochastic problems, the method presented in \cite{liu2022adaptive} and \cite{cai2022selfadaptive} solves for when and how to grow neural networks based on the network error, but it is not easily extensible to problems in which loss cannot be driven to zero. Most similar to the work presented here, AdaNet \cite{cortes2017adanet} approach iteratively grows a neural network before training. It uses the Rademacher complexity to approximate the generalization error of neural network and trains the network predicted to give the least generalization error.

The method presented has similar growth mechanisms to AdaNet, but the method here increases the network size during training by focusing on the network's Mean Squared Error (MSE). Instead of approximating the generalization error as in AdaNet, we train a residual network to approximate the loss of the base neural network. If the residual network can predict structure in the loss of the base network, the algorithm infers that some amount of network complexity is missing from the base neural network, and we combine the two networks into one and continue training. The algorithm then initializes a new residual network and fits the error of the base network again.

\section{Method}

\begin{figure}
\includegraphics[scale=0.4]{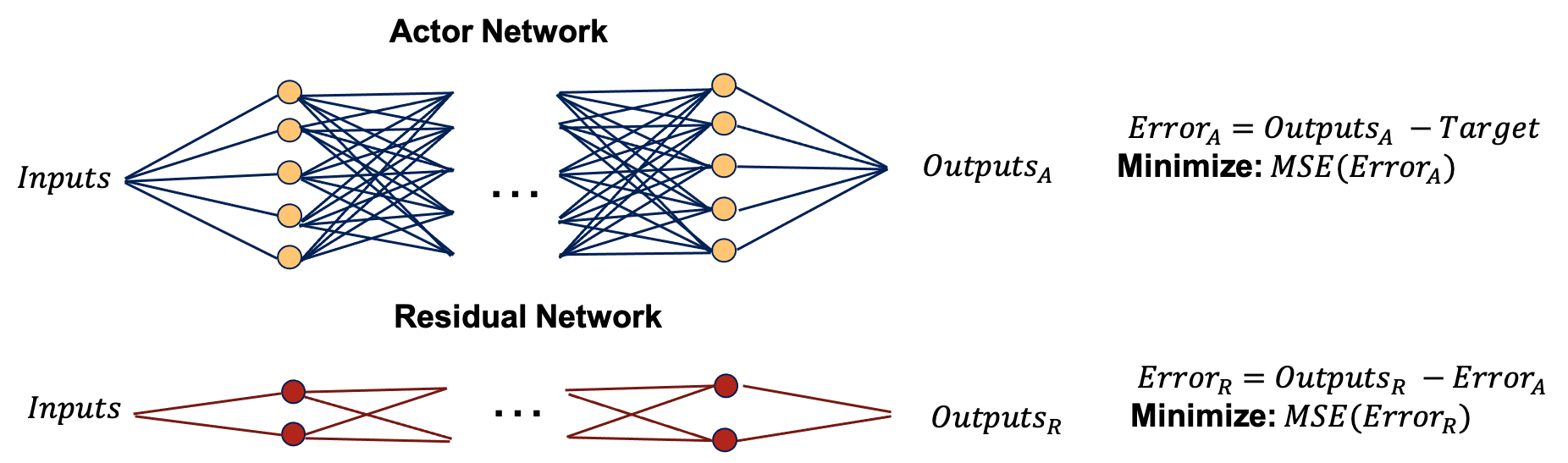}
\caption{Base Network and Residual Network Diagram}
\label{fig:SANN}
\end{figure}

When training a fully connected, multi-layer perceptron with $n$ hidden layers, we initialize a narrower neural network that also has $n$ hidden layers. We call this network the residual neural network. After each training epoch, we train the residual network to fit the current residuals of the larger network (Fig \ref{fig:SANN}).

If adding in the predicted residuals to the larger network’s predictions improves the MSE above a certain user-defined threshold, we increase the latent size of the neural networks by fusing the residual network to the base neural network together. The two, joined networks create a new, fully connected neural network with hidden-layer widths equal to the number of nodes in the base network’s hidden layer plus the number of nodes in the residual network’s hidden layer.

The weights of the first layer and the output layer are completely defined by the base neural network and the residual network. However, the internal, hidden layers have new connections between the base nodes and the residual network’s nodes. These weights need to be initialized. Initializing these weights to zero leads to the smallest MSE after growth, but the backpropagation algorithm will cause these weights to remain equal to each other. Here, we initialize the new weights randomly to have a one tenth the magnitude to the existing weights in the residual network.

After fusing the networks together, we continue to train the larger network. Additionally, we reset the weights of the residual neural network and continue to fit the residual network to the larger network’s residuals. This algorithm is detailed in Algorithm \ref{alg:sann}.

Since this technique focuses on fitting predictable error in the residuals of the network, it is prone to overfitting. One technique to reduce overfitting and the subsequent growth of the neural networks is to use dropout layers during training in both the base neural network and the residual network. In some experiments, networks with dropout perform worse than networks without. So we don't need to rely on dropout to stop excessive network growth, we include a growth criterion that the current mean squared error must be over the growth threshold from the last time the network grew. So if the network last grew at a MSE of $10$, and the growth threshold is $10\%$, the base network must achieve an MSE of $9$ before it can grow again.

\begin{algorithm}
\caption{SANN Algorithm}\label{alg:sann}
\begin{algorithmic}
\State{Training data: $\{X_i\}$, $\{Y_i\}$, adaptation threshold $\gamma$}
\State $f(x)$ Neural Network
\State $g(x)$ Residual Network
\State $\alpha_{prev} \gets 1\times 10^6$ \Comment{set $\alpha_{prev}$ to some large number}
\For{$i$ in number of epochs}
\State Fit $f(x)$ to training data $\{X_i\}$, $\{Y_i\}$. Collect residuals $\{r_i\}$
\State Fit $g(x)$ to $\{X_i\}$, $\{r_i\}$
\State $\alpha \gets \mbox{MSE}(f(x))$
\State $\beta \gets \mbox{MSE}(f(x) + g(x))$
\If{$\beta/\alpha < 1 - \gamma$ and $\alpha/\alpha_{prev} < 1 - \gamma$}
    \State Grow network by combining weights of $f(x)$ and $g(x)$
    \State Reset weights of $g(x)$
    \State $\alpha_{prev} = \alpha$
\EndIf
\EndFor
\end{algorithmic}
\end{algorithm}

\section{Results}

\subsection{CIFAR}
Following AdaNet \cite{cortes2017adanet}, we compare the performance of these networks on CIFAR-10 classification using histogram data from the CIFAR images. The state-of-the art method for image classification typically include convolutional neural networks, but following \cite{cortes2017adanet}, we use histograms of the image data to test the viability of this method. In these CIFAR-10 experiments, we only use histograms of each color distribution, with 40 bins per color, 120 bins total. We evaluate this method on a hold-out set. We test the performance of small fixed networks, small growing networks, large fixed networks, and large growing networks. We test the performance of the large growing networks to ensure that growth does not always occur at all network sizes. We run $10$ training runs for each pairwise classification experiment for each network type.

\begin{figure}
\subfloat[Latent Size]{\includegraphics[scale=0.23]{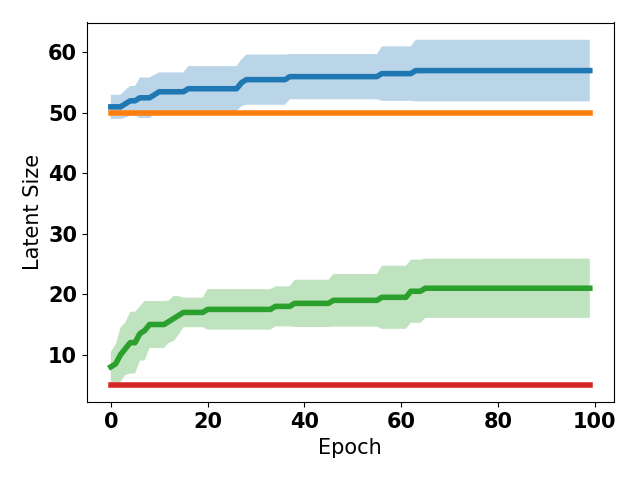}}
\subfloat[MSE]{\includegraphics[scale=0.23]{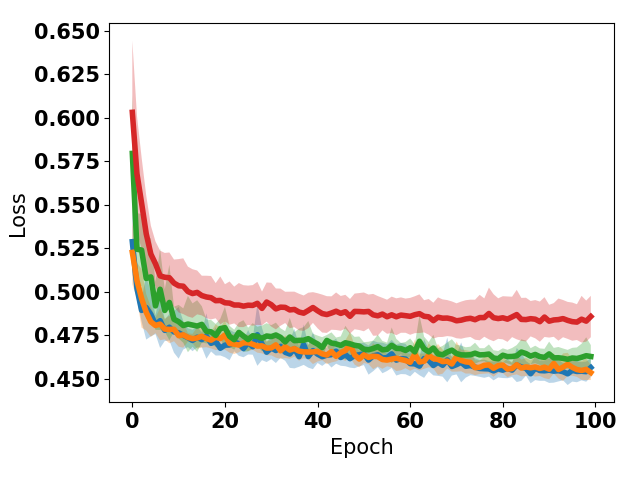}}
\subfloat[Percent Correct]{\includegraphics[scale=0.23]{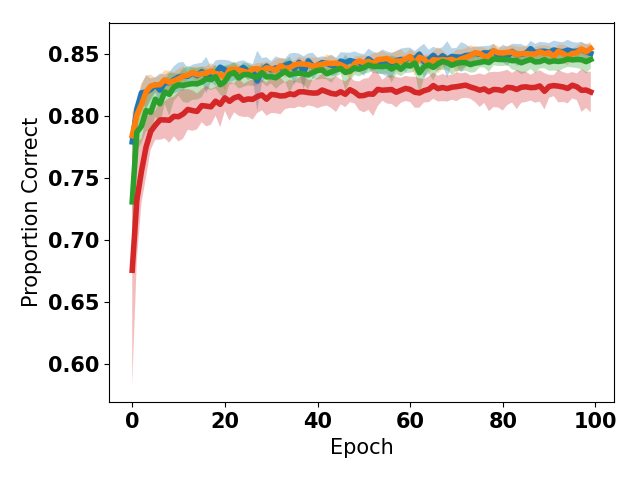}}
\caption{Deer vs Truck}
\label{fig:cifar-deer-truck}
\end{figure}

For the classification of deer vs truck, the growing network performs better than the small, fixed network and similarly to the large networks (Figure \ref{fig:cifar-deer-truck}). The large, growing network grows in size a small amount, but maintains a similar mean squared error to the large, fixed network.

\begin{figure}
\subfloat[Latent Size]{\includegraphics[scale=0.23]{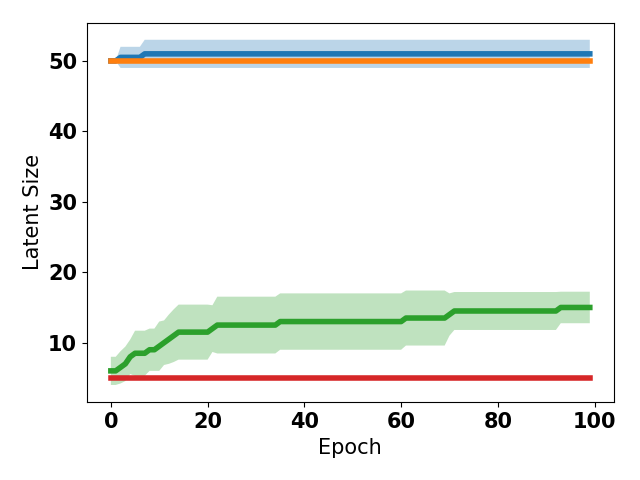}}
\subfloat[MSE]{\includegraphics[scale=0.23]{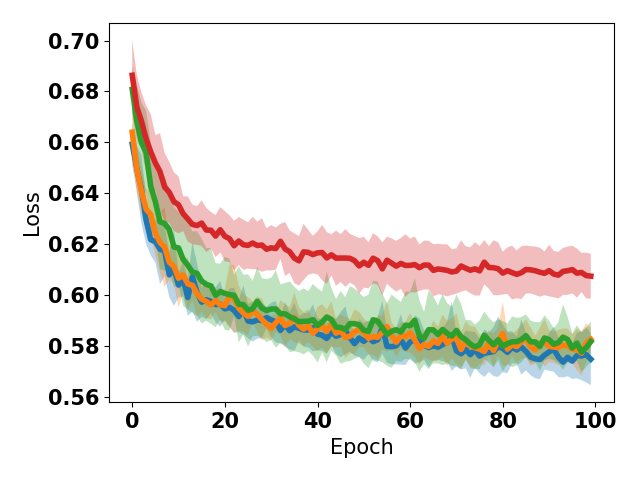}}
\subfloat[Percent Correct]{\includegraphics[scale=0.23]{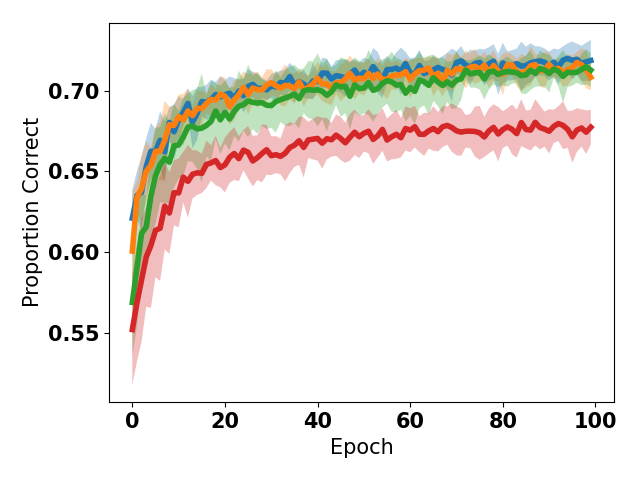}}
\caption{Bird vs Frog}
\label{fig:cifar-bird-frog}
\end{figure}

In testing the networks' performance on the bird vs frog classification problem, we see that the training does not induce as much growth as that seen in the deer vs truck problem (Figure \ref{fig:cifar-bird-frog}). The growing network, again, performs better than the small, fixed network and similarly to the larger networks.

\begin{figure}
\subfloat[Latent Size]{\includegraphics[scale=0.23]{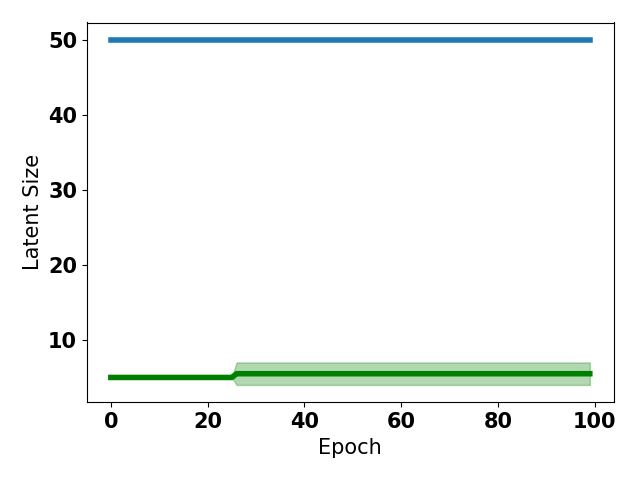}}
\subfloat[MSE]{\includegraphics[scale=0.23]{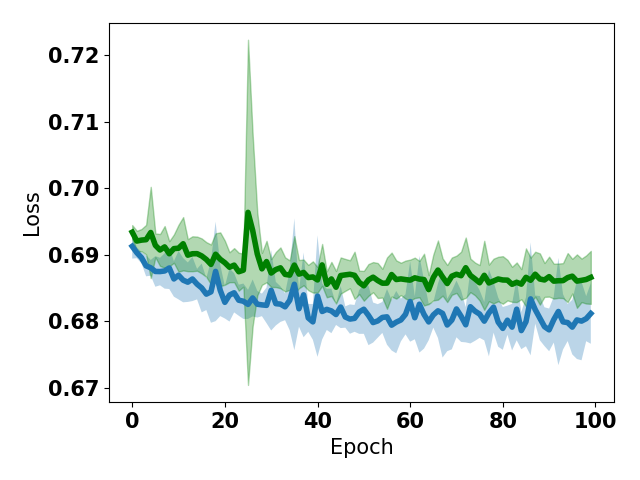}}
\subfloat[Percent Correct]{\includegraphics[scale=0.23]{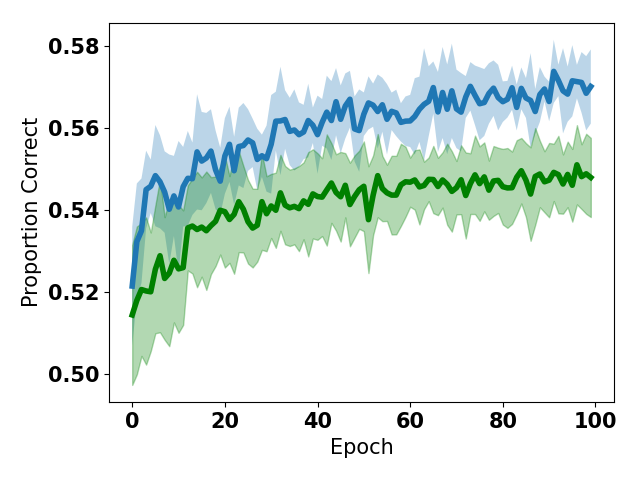}}
\caption{Cat vs Dog}
\label{fig:cifar-cat-dog}
\end{figure}

For the cat vs dog classification problem, neither the small nor large networks typically grow (Figure \ref{fig:cifar-cat-dog}). This lack of growth means that the residual network typically does not find large error in the loss that the base network is not fitting. The smaller network performs worse than the larger network, but this difference is relatively small (less than $0.01$ in mean loss). It is possible that the color histograms for cat vs dog classification do not contain enough information for the neural network to make complex classification decisions. In this case, the neural network for both the small and large networks do not grow.

\subsection{Imitation Learning}

We use imitation learning as another testing task for the adaptive neural network method. We compare performance of networks of varying sizes within the DAgger algorithm \cite{ross2011reduction}. DAgger is a popular method of imitation learning when one has access to the environment and to the expert itself during training. We run the DAgger algorithm to imitate an expert within the PyUXV environment, also used in \cite{9636797}. In this environment, the agent must navigate around obstacles to reach a goal. Here, we imitate the commands needed to avoid the obstacles.

\begin{figure}
\subfloat[Latent Size]{\includegraphics[scale=0.23]{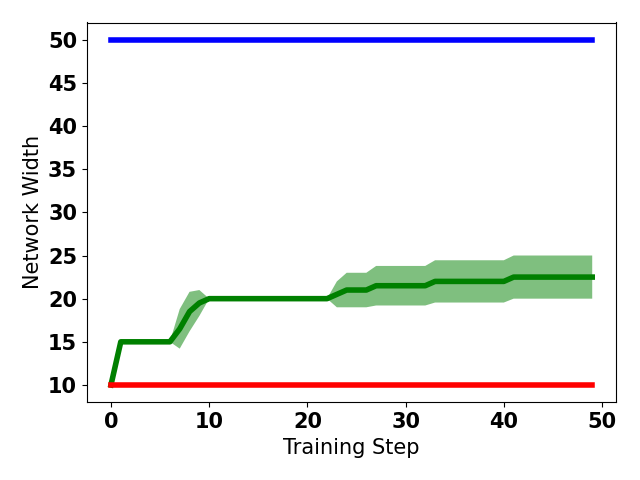}}
\subfloat[MSE]{\includegraphics[scale=0.23]{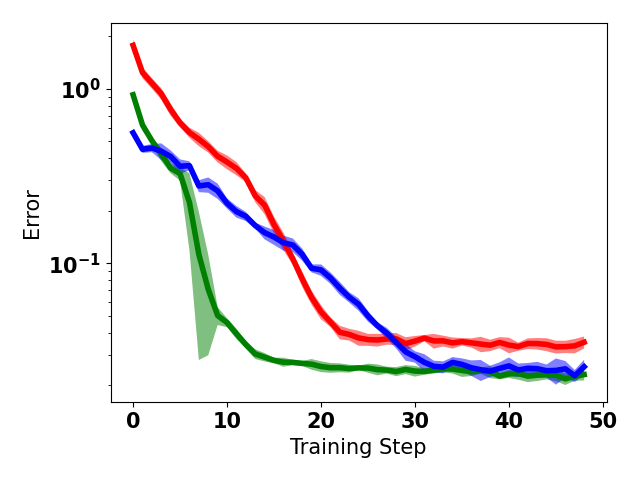}}
\subfloat[Score]{\includegraphics[scale=0.23]{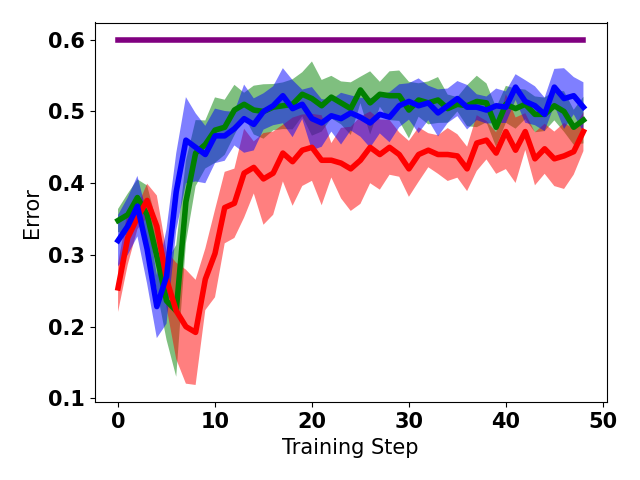}}
\caption{PyUXV with DAgger}
\label{fig:pyuxv}
\end{figure}

With 10 runs in each condition, we find that the growing network trains faster than the large and small fixed networks Figure \ref{fig:pyuxv}. Also, the growing network achieves a similar score to the large network, which is above the score of the small network.

We use Mujoco as a baseline for testing the adaptive networks performance in behavior cloning. We consider two environments: Ant and Half Cheetah, and perform 10 iterations for each conditions in the two environments. We train on 10 expert trajectories and then validate against 10 unseen trajectories. The expert trajectories were generated using \cite{ho2016generative}. We do not use dropout in these experiments as we found that dropout reduced the performance of the networks. Instead, we prevent runaway growth by preventing a network from growing again until it achieves its MSE improvement threshold.

In both Ant and Half Cheetah, the MSE of the growing network jumps at certain moments during training. These jumps occur right after the network grows as the evaluation occurs before the network’s new, randomly initialized weights have been trained.

\begin{figure}
\subfloat[Latent Size]{\includegraphics[scale=0.23]{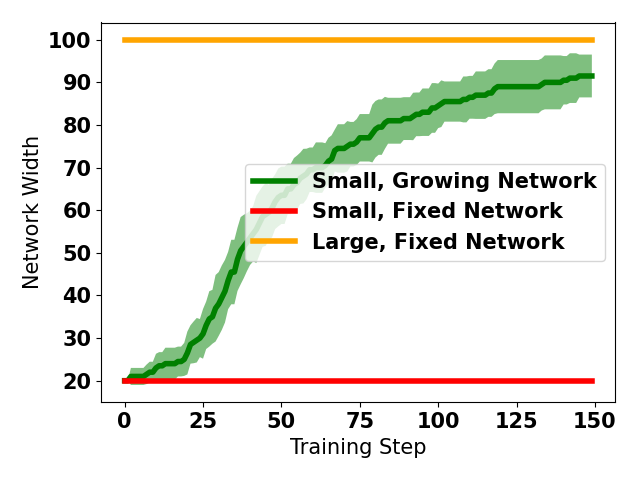}}
\subfloat[MSE]{\includegraphics[scale=0.23]{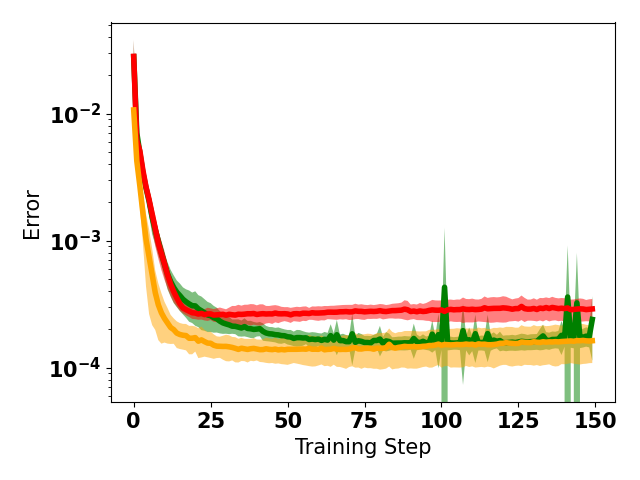}}
\subfloat[Score]{\includegraphics[scale=0.23]{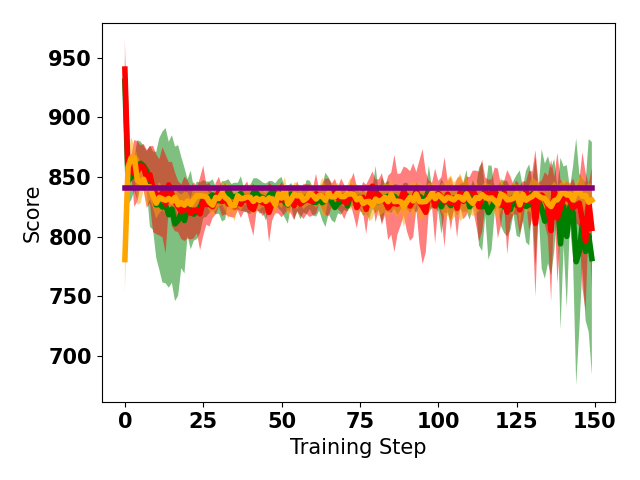}}
\caption{Behavior Cloning on Ant}
\label{fig:ant}
\end{figure}

For the Ant environment, the growing network provides gains in MSE to the small, fixed network and a similar MSE to the large, growing network (Figure \ref{fig:ant}). The three networks attain a similar score in the environment, but the large-growing error has less variance in the score.

\begin{figure}
\subfloat[Latent Size]{\includegraphics[scale=0.23]{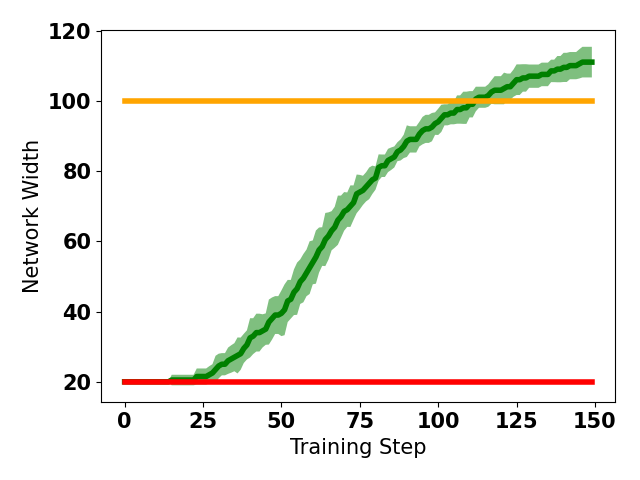}}
\subfloat[MSE]{\includegraphics[scale=0.23]{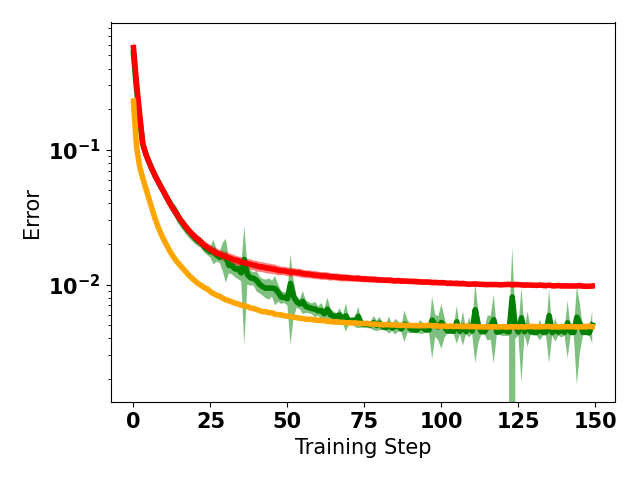}}
\subfloat[Score]{\includegraphics[scale=0.23]{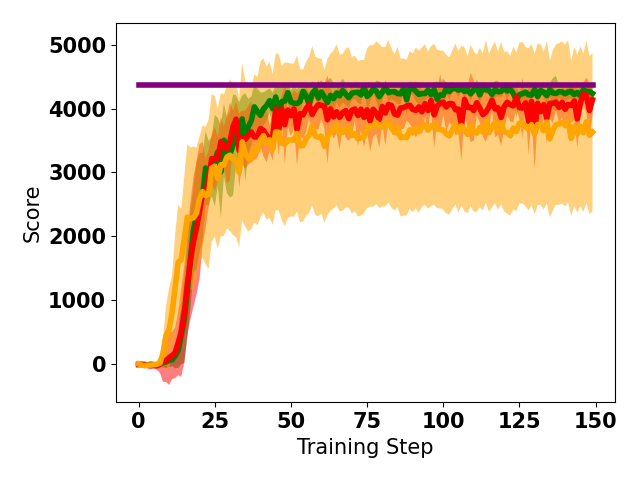}}
\caption{Behavior Cloning on Half Cheetah}
\label{fig:cheetah}
\end{figure}

In the Half Cheetah environment, the network on average grows to be larger than the large, fixed network. However, before reaching the larger size, the growing network achieves a similar MSE to the larger network (ignoring the steps in which one of the network grows and spikes the MSE). The imitators with growing network appears to achieve an improved MSE to the large, fixed network (Figure \ref{fig:cheetah}). Also, imitators with small, growing network achieve the highest average environmental rewards.

\section{Adaptive Networks in Reinforcement Learning}

We apply the same adaptive-network technique to the value function within the reinforcement-learning algorithm PPO (\cite{DBLP:journals/corr/SchulmanWDRK17}). We grow the size of the value-function network if its fit can be improved by the addition of the residual network. In all the experimental conditions, the policy network is fixed to have two hidden layers that are each 64 nodes in width while the value function has two hidden layers with varying widths. We do not use dropout in these experiments, and without dropout, the networks can experience continuing growth, even for large networks.

\begin{figure}
\centering
\subfloat[Latent Size]{\includegraphics[scale=0.3]{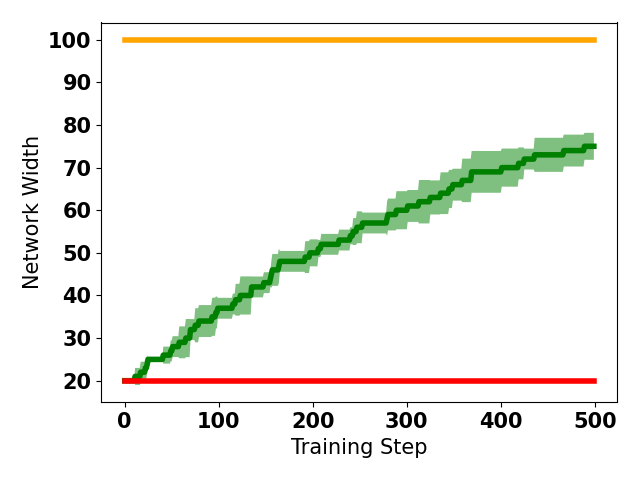}}
\subfloat[Score]{\includegraphics[scale=0.3]{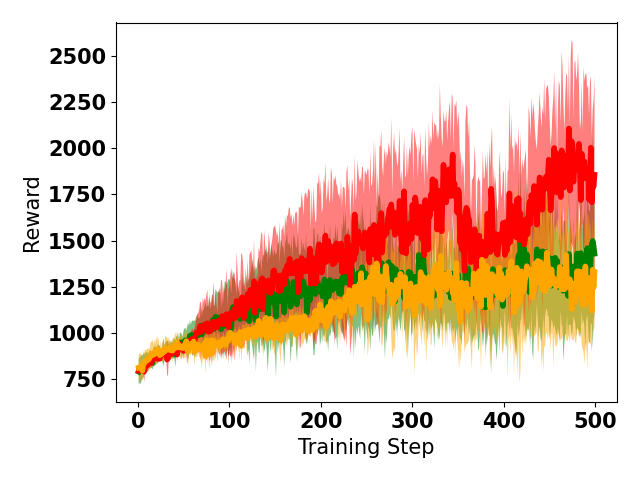}}
\caption{Reinforcement Learning on Ant}
\label{fig:antrl}
\end{figure}

\begin{figure}
\centering
\subfloat[Latent Size]{\includegraphics[scale=0.3]{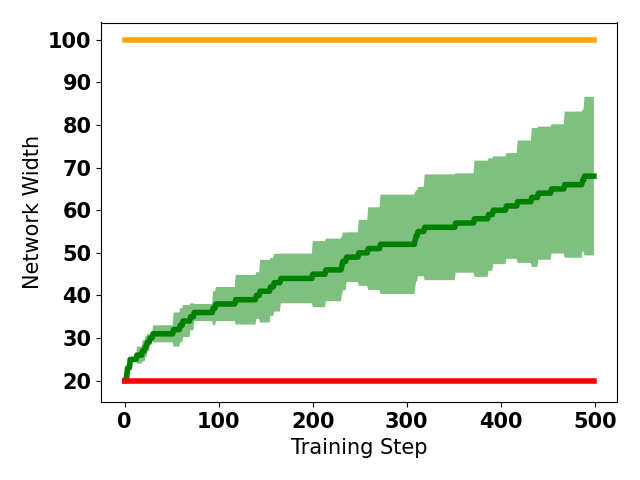}}
\subfloat[Score]{\includegraphics[scale=0.3]{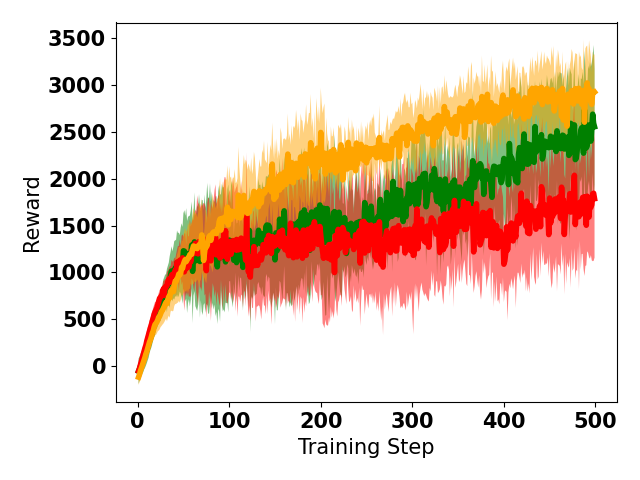}}
\caption{Reinforcement Learning on Cheetah}
\label{fig:cheetahrl}
\end{figure}

We train using PPO for the varying value networks in both the Mujoco Ant and Half Cheetah Environment. We trained five runs for each condition in each environment. In the Ant environment, we see that the growing network performs similarly to the large network (Fig \ref{fig:antrl}). However, the PPO runs with a smaller value network outperforms the runs with both the larger and growing value networks. In the Half Cheetah environment, the PPO runs with the growing network begin by performing similarly to the small network, but as the value network grows, the score for these runs begin to approach the score of the PPO runs with the large value network (Fig \ref{fig:cheetahrl}).

It appears that the PPO runs with growing value networks can perform similarly to the PPO runs with larger networks, but may take more time reach a similar proficiency. More work is required to further explore the effectiveness of these growing value networks and to reduce the unnecessary growing of these networks. 

\section{Conclusion}
Here we studied the technique of fitting a small neural network to the residuals of another neural network and combining those networks if the residual network finds significant, predictable error in the base network’s fit. Using test cases from classification, imitation learning, and reinforcement learning, we find that this technique can produce effective networks without having preexisting knowledge of the problem's complexity.

In many cases this method of fitting residuals can achieve evaluation losses that are similar to much larger multi-layer perceptrons for several types of problems. In fitting CIFAR data, we find that the networks would grow to different sizes for different problems, which may reflect the available information in the histogram data. In imitation learning, we find that the adaptive networks used for behavioral cloning and within the DAgger algorithm can achieve similar effectiveness to networks of fixed size that began with much wider hidden layers. The reinforcement-learning runs using an adaptive value networks appear to train more slowly that the larger, fixed networks, but can also achieve similar results.

Future work will focus on improving the initialization of new weights during the network combination process using methods like \cite{evci2022gradmax}. We would also like to implement growth with varying network depth like in AdaNet (\cite{cortes2017adanet}).

\section*{Acknowledgement}
This research is in part funded by the Test Resource Management Center (TRMC) and Test Evaluation/Science \& Technology (T\&E/S\&T) Program under contract W900KK-19-C-004.

Any opinions, findings and conclusions or recommendations expressed in this material are those of the author(s) and do not necessarily reflect the views of the Test Resource Management Center (TRMC) and Test Evaluation/Science \& Technology (T\&E/S\&T) Program.

Distribution Statement A - Approved for public release.

\bibliography{bibliography}

\begin{thebibliography}{15}
\expandafter\ifx\csname natexlab\endcsname\relax\def\natexlab#1{#1}\fi
\expandafter\ifx\csname bibnamefont\endcsname\relax
  \def\bibnamefont#1{#1}\fi
\expandafter\ifx\csname bibfnamefont\endcsname\relax
  \def\bibfnamefont#1{#1}\fi
\expandafter\ifx\csname citenamefont\endcsname\relax
  \def\citenamefont#1{#1}\fi
\expandafter\ifx\csname url\endcsname\relax
  \def\url#1{\texttt{#1}}\fi
\expandafter\ifx\csname urlprefix\endcsname\relax\def\urlprefix{URL }\fi
\providecommand{\bibinfo}[2]{#2}
\providecommand{\eprint}[2][]{\url{#2}}

\bibitem[{\citenamefont{Frankle and Carbin}(2019)}]{frankle2019lottery}
\bibinfo{author}{\bibfnamefont{J.}~\bibnamefont{Frankle}} \bibnamefont{and}
  \bibinfo{author}{\bibfnamefont{M.}~\bibnamefont{Carbin}},
  \emph{\bibinfo{title}{The lottery ticket hypothesis: Finding sparse,
  trainable neural networks}} (\bibinfo{year}{2019}), \eprint{1803.03635}.

\bibitem[{\citenamefont{Elsken et~al.}(2019)\citenamefont{Elsken, Metzen, and
  Hutter}}]{elsken2019neural}
\bibinfo{author}{\bibfnamefont{T.}~\bibnamefont{Elsken}},
  \bibinfo{author}{\bibfnamefont{J.~H.} \bibnamefont{Metzen}},
  \bibnamefont{and} \bibinfo{author}{\bibfnamefont{F.}~\bibnamefont{Hutter}},
  \emph{\bibinfo{title}{Neural architecture search: A survey}}
  (\bibinfo{year}{2019}), \eprint{1808.05377}.

\bibitem[{\citenamefont{Blalock et~al.}(2020)\citenamefont{Blalock, Ortiz,
  Frankle, and Guttag}}]{blalock2020state}
\bibinfo{author}{\bibfnamefont{D.}~\bibnamefont{Blalock}},
  \bibinfo{author}{\bibfnamefont{J.~J.~G.} \bibnamefont{Ortiz}},
  \bibinfo{author}{\bibfnamefont{J.}~\bibnamefont{Frankle}}, \bibnamefont{and}
  \bibinfo{author}{\bibfnamefont{J.}~\bibnamefont{Guttag}},
  \emph{\bibinfo{title}{What is the state of neural network pruning?}}
  (\bibinfo{year}{2020}), \eprint{2003.03033}.

\bibitem[{\citenamefont{Alparslan et~al.}(2021)\citenamefont{Alparslan, Moyer,
  Isozaki, Schwartz, Dunlop, Dave, and Kim}}]{alparslan2021searching}
\bibinfo{author}{\bibfnamefont{Y.}~\bibnamefont{Alparslan}},
  \bibinfo{author}{\bibfnamefont{E.~J.} \bibnamefont{Moyer}},
  \bibinfo{author}{\bibfnamefont{I.~M.} \bibnamefont{Isozaki}},
  \bibinfo{author}{\bibfnamefont{D.}~\bibnamefont{Schwartz}},
  \bibinfo{author}{\bibfnamefont{A.}~\bibnamefont{Dunlop}},
  \bibinfo{author}{\bibfnamefont{S.}~\bibnamefont{Dave}}, \bibnamefont{and}
  \bibinfo{author}{\bibfnamefont{E.}~\bibnamefont{Kim}},
  \emph{\bibinfo{title}{Towards searching efficient and accurate neural network
  architectures in binary classification problems}} (\bibinfo{year}{2021}),
  \eprint{2101.06511}.

\bibitem[{\citenamefont{So et~al.}(2019)\citenamefont{So, Liang, and
  Le}}]{so2019evolved}
\bibinfo{author}{\bibfnamefont{D.~R.} \bibnamefont{So}},
  \bibinfo{author}{\bibfnamefont{C.}~\bibnamefont{Liang}}, \bibnamefont{and}
  \bibinfo{author}{\bibfnamefont{Q.~V.} \bibnamefont{Le}},
  \emph{\bibinfo{title}{The evolved transformer}} (\bibinfo{year}{2019}),
  \eprint{1901.11117}.

\bibitem[{\citenamefont{Molchanov et~al.}(2017)\citenamefont{Molchanov, Tyree,
  Karras, Aila, and Kautz}}]{molchanov2017pruning}
\bibinfo{author}{\bibfnamefont{P.}~\bibnamefont{Molchanov}},
  \bibinfo{author}{\bibfnamefont{S.}~\bibnamefont{Tyree}},
  \bibinfo{author}{\bibfnamefont{T.}~\bibnamefont{Karras}},
  \bibinfo{author}{\bibfnamefont{T.}~\bibnamefont{Aila}}, \bibnamefont{and}
  \bibinfo{author}{\bibfnamefont{J.}~\bibnamefont{Kautz}},
  \emph{\bibinfo{title}{Pruning convolutional neural networks for resource
  efficient inference}} (\bibinfo{year}{2017}), \eprint{1611.06440}.

\bibitem[{\citenamefont{Kilcher et~al.}(2019)\citenamefont{Kilcher,
  B{\'e}cigneul, and Hofmann}}]{kilcher2019escaping}
\bibinfo{author}{\bibfnamefont{Y.}~\bibnamefont{Kilcher}},
  \bibinfo{author}{\bibfnamefont{G.}~\bibnamefont{B{\'e}cigneul}},
  \bibnamefont{and} \bibinfo{author}{\bibfnamefont{T.}~\bibnamefont{Hofmann}},
  \emph{\bibinfo{title}{Escaping flat areas via function-preserving structural
  network modifications}} (\bibinfo{year}{2019}),
  \urlprefix\url{https://openreview.net/forum?id=H1eadi0cFQ}.

\bibitem[{\citenamefont{Evci et~al.}(2022)\citenamefont{Evci, Vladymyrov,
  Unterthiner, van Merri{\"e}nboer, and Pedregosa}}]{evci2022gradmax}
\bibinfo{author}{\bibfnamefont{U.}~\bibnamefont{Evci}},
  \bibinfo{author}{\bibfnamefont{M.}~\bibnamefont{Vladymyrov}},
  \bibinfo{author}{\bibfnamefont{T.}~\bibnamefont{Unterthiner}},
  \bibinfo{author}{\bibfnamefont{B.}~\bibnamefont{van Merri{\"e}nboer}},
  \bibnamefont{and}
  \bibinfo{author}{\bibfnamefont{F.}~\bibnamefont{Pedregosa}},
  \emph{\bibinfo{title}{Gradmax: Growing neural networks using gradient
  information}} (\bibinfo{year}{2022}), \eprint{2201.05125}.

\bibitem[{\citenamefont{Liu et~al.}(2022)\citenamefont{Liu, Cai, and
  Chen}}]{liu2022adaptive}
\bibinfo{author}{\bibfnamefont{M.}~\bibnamefont{Liu}},
  \bibinfo{author}{\bibfnamefont{Z.}~\bibnamefont{Cai}}, \bibnamefont{and}
  \bibinfo{author}{\bibfnamefont{J.}~\bibnamefont{Chen}},
  \emph{\bibinfo{title}{Adaptive two-layer relu neural network: I. best
  least-squares approximation}} (\bibinfo{year}{2022}), \eprint{2107.08935}.

\bibitem[{\citenamefont{Cai et~al.}(2022)\citenamefont{Cai, Chen, and
  Liu}}]{cai2022selfadaptive}
\bibinfo{author}{\bibfnamefont{Z.}~\bibnamefont{Cai}},
  \bibinfo{author}{\bibfnamefont{J.}~\bibnamefont{Chen}}, \bibnamefont{and}
  \bibinfo{author}{\bibfnamefont{M.}~\bibnamefont{Liu}},
  \emph{\bibinfo{title}{Self-adaptive deep neural network: Numerical
  approximation to functions and pdes}} (\bibinfo{year}{2022}),
  \eprint{2109.02839}.

\bibitem[{\citenamefont{Cortes et~al.}(2017)\citenamefont{Cortes, Gonzalvo,
  Kuznetsov, Mohri, and Yang}}]{cortes2017adanet}
\bibinfo{author}{\bibfnamefont{C.}~\bibnamefont{Cortes}},
  \bibinfo{author}{\bibfnamefont{X.}~\bibnamefont{Gonzalvo}},
  \bibinfo{author}{\bibfnamefont{V.}~\bibnamefont{Kuznetsov}},
  \bibinfo{author}{\bibfnamefont{M.}~\bibnamefont{Mohri}}, \bibnamefont{and}
  \bibinfo{author}{\bibfnamefont{S.}~\bibnamefont{Yang}},
  \emph{\bibinfo{title}{Adanet: Adaptive structural learning of artificial
  neural networks}} (\bibinfo{year}{2017}), \eprint{1607.01097}.

\bibitem[{\citenamefont{Ross et~al.}(2011)\citenamefont{Ross, Gordon, and
  Bagnell}}]{ross2011reduction}
\bibinfo{author}{\bibfnamefont{S.}~\bibnamefont{Ross}},
  \bibinfo{author}{\bibfnamefont{G.~J.} \bibnamefont{Gordon}},
  \bibnamefont{and} \bibinfo{author}{\bibfnamefont{J.~A.}
  \bibnamefont{Bagnell}}, \emph{\bibinfo{title}{A reduction of imitation
  learning and structured prediction to no-regret online learning}}
  (\bibinfo{year}{2011}), \eprint{1011.0686}.

\bibitem[{\citenamefont{Lowman et~al.}(2021)\citenamefont{Lowman, McClellan,
  and Mullins}}]{9636797}
\bibinfo{author}{\bibfnamefont{C.~A.} \bibnamefont{Lowman}},
  \bibinfo{author}{\bibfnamefont{J.~S.} \bibnamefont{McClellan}},
  \bibnamefont{and} \bibinfo{author}{\bibfnamefont{G.~E.}
  \bibnamefont{Mullins}}, in \emph{\bibinfo{booktitle}{2021 IEEE/RSJ
  International Conference on Intelligent Robots and Systems (IROS)}}
  (\bibinfo{year}{2021}), pp. \bibinfo{pages}{8921--8927}.

\bibitem[{\citenamefont{Ho and Ermon}(2016)}]{ho2016generative}
\bibinfo{author}{\bibfnamefont{J.}~\bibnamefont{Ho}} \bibnamefont{and}
  \bibinfo{author}{\bibfnamefont{S.}~\bibnamefont{Ermon}},
  \emph{\bibinfo{title}{Generative adversarial imitation learning}}
  (\bibinfo{year}{2016}), \eprint{1606.03476}.

\bibitem[{\citenamefont{Schulman et~al.}(2017)\citenamefont{Schulman, Wolski,
  Dhariwal, Radford, and Klimov}}]{DBLP:journals/corr/SchulmanWDRK17}
\bibinfo{author}{\bibfnamefont{J.}~\bibnamefont{Schulman}},
  \bibinfo{author}{\bibfnamefont{F.}~\bibnamefont{Wolski}},
  \bibinfo{author}{\bibfnamefont{P.}~\bibnamefont{Dhariwal}},
  \bibinfo{author}{\bibfnamefont{A.}~\bibnamefont{Radford}}, \bibnamefont{and}
  \bibinfo{author}{\bibfnamefont{O.}~\bibnamefont{Klimov}},
  \bibinfo{journal}{CoRR} \textbf{\bibinfo{volume}{abs/1707.06347}}
  (\bibinfo{year}{2017}), \eprint{1707.06347},
  \urlprefix\url{http://arxiv.org/abs/1707.06347}.

\end{thebibliography}
\end{document}